\documentclass[conference]{IEEEtran}
\IEEEoverridecommandlockouts
\usepackage{cite}
\usepackage{amsmath,amssymb,amsfonts}
\usepackage{algorithmic}
\usepackage{graphicx}
\usepackage{textcomp}

\usepackage[dvipsnames]{xcolor}
\usepackage[acronym]{glossaries}
\usepackage{xspace}
\usepackage{array}
\usepackage{tabularx}
\usepackage{multirow}
\usepackage{multirow}

\usepackage{array}
\usepackage{color}
\usepackage{colortbl}
\usepackage{hhline}

\begin{document}

\title{
Accelerate Model Parallel Training by Using Efficient Graph Traversal Order in Device Placement

\thanks{This work was supported by the ExtremeEarth project funded by European Union’s Horizon 2020 Research and Innovation Programme under Grant agreement no. 825258.

Preprint. Under review.}
}

\author{\IEEEauthorblockN{Tianze Wang,  Amir H. Payberah, Desta Haileselassie Hagos, and Vladimir Vlassov}
\IEEEauthorblockA{
\textit{KTH Royal Institute of Technology}, Stockholm, Sweden\\
tianzew@kth.se, payberah@kth.se, destah@kth.se, vladv@kth.se\\
}}

\maketitle

\thispagestyle{plain}
\pagestyle{plain}

\begin{abstract}
Modern neural networks require long training to reach decent performance on massive datasets. One common approach to speed up training is model parallelization, where large neural networks are split across multiple devices. However, different device placements of the same neural network lead to different training times. Most of the existing device placement solutions treat the problem as sequential decision-making by traversing neural network graphs and assigning their neurons to different devices. This work studies the impact of graph traversal order on device placement. In particular, we empirically study how different graph traversal order leads to different device placement, which in turn affects the training execution time. Our experiment results show that the best graph traversal order depends on the type of neural networks and their computation graphs features. In this work, we also provide recommendations on choosing graph traversal order in device placement for various neural network families to improve the training time in model parallelization.
\end{abstract}

\begin{IEEEkeywords}
Directed Acyclic Graph, Graph 
Traversal Order, Device Placement, Deep Learning
\end{IEEEkeywords}

\newacronym{nmt}{NMT}{Neural Machine Translation}
\newacronym{DL}{DL}{Deep Learning}
\newacronym{GNN}{GNN}{Graph Neural Network}
\newacronym{DAG}{DAG}{Directed Acyclic Graph}
\newacronym{RL}{RL}{Reinforcement Learning}
\newacronym{MDP}{MDP}{Markov Decision Process}
\newacronym{DFS}{DFS}{Depth-First Search}
\newacronym{BFS}{BFS}{Breadth-First Search}
\newacronym{NAS}{NAS}{Neural Architecture Search}
\newacronym{EO}{EO}{Earth Observation}
\newacronym{RS}{RS}{Remote Sensing}
\newacronym{SAR}{SAR}{Synthetic Aperture Radar}

\newcommand{\nmt}{\texttt{nmt}\xspace}
\newcommand{\ptb}{\texttt{ptb}\xspace}
\newcommand{\cifar}{\texttt{cifar10}\xspace}

\newcommand{\topo}{\texttt{topo}\xspace}
\newcommand{\reversedtopo}{\texttt{reversed-topo}\xspace}
\newcommand{\dfspre}{\texttt{dfs-preorder}\xspace}
\newcommand{\dfspost}{\texttt{dfs-postorder}\xspace}
\newcommand{\bfs}{\texttt{bfs}\xspace}
\newcommand{\lexi}{\texttt{lexico}\xspace}

\section{Introduction}
Recent years have seen the prevalence of \gls{DL} with larger and deeper models with billions of neurons~\cite{shoeybi2019megatron, brown2020language}. Together with the performance boost of \gls{DL} models comes the increasing computation demand for model training. Most solutions seek to parallelize the training on GPU clusters to meet the requirement of computation power. {\em Data parallelism}~\cite{shallue2019measuring} and {\em model parallelism}~\cite{shoeybi2019megatron} of \gls{DL} models are the most common parallelization strategies. In data parallelism, data are distributed among several servers (a.k.a. workers or devices) in a GPU cluster. In contrast, in model parallelism, the \gls{DL} model is split into multiple parts and distributed among workers. Assigning different parts of a \gls{DL} model to different workers is known as {\em device placement}.

Finding the optimal device placement of \gls{DL} models in model parallelization is challenging. It is mainly due to the large search spaces of potential parallelization strategies, understanding model architectures, and device characteristics~\cite{zhou2019gdp}. Despite lots of efforts to improve device placements, the training time of device placement methods is still very long~\cite{mirhoseini2018hierarchical, addanki2019placeto, zhou2019gdp, gao2018post, gao2018spotlight}. The first effort in automating device placement combines global partitioning and local scheduling by using heuristic strategies to first partition the \gls{DL} model into smaller parts and then determine the execution schedule of neurons within each part~\cite{mayer2017tensorflow}. 

The state-of-the-art device placement methods use a combination of \gls{GNN} and \gls{RL} to find the placement of \gls{DL} models~\cite{addanki2019placeto, zhou2019gdp}. In these solutions, the computation graph of a \gls{DL} model is represented as a \gls{DAG}, in which each node of the \gls{DAG} represents a single operation or a group of operations. In a typical setting, a \gls{GNN} takes a \gls{DAG} of a \gls{DL} model and its nodes' features as input and generates nodes embeddings, which summarize the attributes and neighborhood topology of each node~\cite{addanki2019placeto, zhou2019gdp}. An \gls{RL} agent then processes the node embeddings and uses a policy to predict device placements for all nodes in the \gls{DAG} on the given device cluster. To this end, the \gls{RL} agent needs to traverse all the nodes in the \gls{DAG} and learn to propose placements to reduce the training time of the \gls{DL}.

Identifying a good graph traversal order can decrease the \gls{RL} agent training time and can also potentially help the \gls{RL} agent to find better placements to reduce the \gls{DL} execution time. In this work, we empirically study the relationship between graph traversal order in device placement and the learning efficiency of the \gls{RL} agent for device placement during the training process. We look into six different graph traversal order and show how they affect the training process of Placeto~\cite{addanki2019placeto}, a state-of-the-art device placement method on three different families of neural networks. Each family of neural networks contains structurally similar \gls{DL} models~\cite{pham2018efficient}. Our initial results suggest that different traversal order are better suited for different types of neural networks. The best graph traversal order to use also depends on the attributes of the \gls{DL} model that we want to find suitable placements on.

We also explain how our methods on some models could be used in \gls{RS} and \gls{EO}. \gls{RS} and \gls{EO} are domains where there is a need to provide near real-time services and products for global monitoring of planet earth. \gls{EO} satellites developed over the years have provided an unprecedented amount of data that need to be processed~\cite{hagos2021extremeearth,zhu2017deep}. Model parallelization methods can contribute to these domains by distributing the computation and memory requirement for training large models on large datasets. 

Our contribution is summarized as follows:
\begin{enumerate}
    \item We empirically study the impact of the graph traversal order on finding the best device placement for a model-parallel distributed \gls{DL} model and, as a consequence, on the training time of the distributed \gls{DL} model. In this study, we considered different architectures of \gls{DL} models, namely, CNN and RNN. Our study shows that different graph traversal order triumphs at finding the best device placements efficiently for different types of \gls{DL} models.
    \item Based on our empirical evaluation of graph traversal order in device placement for different model-parallel \gls{DL} architectures, we summarize and provide guidelines on identifying the best graph traversal order for a given \gls{DL} model based on its characteristics. For example, we recommend using \bfs traversal order for model-parallel RNNs with large average degrees to perform device placement.
    \item In the context of \gls{RS} and \gls{EO}, we show how our methods on identifying the best graph traversal order can be used on the \gls{DL} models in the Polar Use case (e.g., CNN models for satellite image classification) and in the Food Security Use case (e.g., an RNN model for sequence classification). 
    Our work on device placement allows finding the best placements for \gls{DL} models faster than not identifying the best graph traversal order to use. 
    The better device placement improves the training performance in model-parallel distributed \gls{DL}. Choosing a proper graph traversal method in device placement (1) improves distributed training time of complex \gls{DL} models, and (2) allows training \gls{DL} models on a larger dataset within a certain (the same) amount of time. The above two-fold benefit enables real-time online training of model-parallel distributed DL models with time deadlines.
\end{enumerate}

\section{Preliminaries}
In this section, we discuss problem formulation of device placement problem, graph embedding, \gls{RL} approach for device placement, and graph traversal order.

\subsection{Device Placement}
Let $G(V, E)$ be a \gls{DAG} that represents the computation graph of a neural network. Each node $v \in V$ describes a single computation operation (e.g., convolution) or a predefined small group of operations (e.g., groups of convolutions nearby) that we are interested in predicting its device placement. Each edge $e \in E$ models the data dependencies between the vertices. Let $D$ denotes a given device cluster (e.g., GPU clusters) where $d \in D$ characterizes a single device in $D$. A placement $p : V \rightarrow D$ is a mapping that assigns each node in $G$ to a device in $D$. Our goal in device placement is to find a placement $p$ to minimize the training time of $G$ (i.e., the \gls{DL} model) on the given device cluster $D$ while satisfying the memory constraints of every device in the cluster.

When given a fixed number of devices, we can treat the device placement task as a classification problem by considering each device identifier as a label. The classification model takes the DAG of a computation graph $G$ as input and classifies every neuron or group of neurons of $G$ into devices in $D$. To this end, Placeto~\cite{addanki2019placeto} models a device placement task as ``finding a sequence of iterative placement improvements.'' In each training round, Placeto takes the current placement for the DAG and the representation of one of its nodes as input and predicts that node's placement in that DAG. Each round of training last until the placements of all the nodes have been updated once. In the rest of this work, we will use the device placement method as proposed in Placeto.

\subsection{Placeto}
Placeto, in general, consists of two parts, (i) using \gls{GNN}~\cite{scarselli2008graph} for making DAG's embedding, and (ii) using \gls{RL} for assigning nodes to devices. Below, we elaborate these two parts in more details.

\subsubsection{Graph Embedding}
What matters in embedding of a computation graph is not only the nodes features but also their relationship. Thus, if two connected nodes are placed on two different devices, there will be data transfer between two devices in both forward and backward path of model training, which is expensive. Things become even more complicated when there are more complex graph and sub-graph structures. Convolution blocks~\cite{szegedy2015going} contain parallel computation threads that depend on the same node for input data and send the result to another node for intermediate result concatenation. The attention mechanism~\cite{vaswani2017attention} uses a weighted sum of the results from previous layers, which can incur a lot of data communication if the nodes in the previous layers are located on different devices. Temporal dependency~\cite{hochreiter1997long} during the training of recurrent neural networks can also incur a lot of data communication if the nodes that construct the recurrent unit are located on different devices.

\gls{GNN}~\cite{scarselli2008graph} can generate d-dimensional graph embeddings for each node in a given graph that can generalize to unseen graphs. Placeto~\cite{addanki2019placeto} uses a graph embedding architecture that computes node attributes (e.g., the execution time of operation, total size of output tensor), summarizes the topology of a local neighborhood through message passing, and uses pooling operations for creating a global summary of the entire graph. Mitropolitsky et al.~\cite{mitropolitsky2020graph} study the impact of different graph embedding techniques on the execution time of the placement and the computation time of graph embedding techniques. By explicitly modeling the relationships between nodes in a computation graph, better placement can be found by auto device placement methods.

\subsubsection{Reinforcement Learning}
After generating graph embedding of the input \gls{DAG}, Placeto uses a two-layer feed-forward neural network that takes the \gls{DAG} graph embeddings as input to iteratively predict the device placement of the \gls{DAG}'s nodes. The output of the neural network is the probability distribution of the current node over candidate hardware devices. The state-of-the-art methods usually evaluate placements of \gls{DL} models by the execution time of the \gls{DL} models. However, the execution time is not differentiable concerning the parameters in the neural network. Thus, Placeto leverages \gls{RL} for training. 

During the training process, the \gls{RL} agent interacts with the training environment and uses execution time as a reward function to guide the training process. Placeto has a simulator that can predict the execution time of a placement, which helps to speed up the training process of \gls{RL} agent by avoiding taking execution time measurement of placements on real hardware. In each episode of training, Placeto updates the placement of each node one time. The \gls{RL} agent of Placeto is trained with REINFORCE~\cite{williams1992simple} policy-gradient method. 

\subsection{Graph Traversal Order}
\label{sec:preliminaries:graph_traversal_order}
Since the device placement problem is treated as a sequential decision-making task~\cite{mirhoseini2018hierarchical}, we need to convert the computation graph into a sequence of nodes. Placeto formulated the device placement problem as \gls{MDP}, where the \gls{RL} agent selects to update the placement for a node in the computation graph in each state. Thus, we need to form a sequence by traversing the computation graph, which is represented as a \gls{DAG}. Below, we review some of the graph traversal order on \gls{DAG} that one can consider using.

\subsubsection{Topological}
Topological ordering~\cite{kahn1962topological} on the \gls{DAG} of a computation graph defines a graph traversal order such that for every directed edge $u\rightarrow v$ from node $u$ to node $v$, $u$ must appear before $v$ in the traversal order. Topological ordering can be used to represent dependencies in a computation graph where we only visit a node once all its dependencies have been met.

\subsubsection{Reversed Topological}
A reversed topological ordering of a DAG of a computation graph is simply the reversed order of its topological ordering.

\subsubsection{Depth-first Search}
\gls{DFS} is a graph traversal method that starts at source nodes (input nodes of the computation graph) and explores the graph as far as possible by continuously visiting the children nodes of the current node first before visiting the sibling nodes. A \gls{DFS} ordering is an enumeration of the nodes that is a possible output of applying \gls{DFS} on the graph. A \gls{DFS} preorder is a list of nodes that are in the order of when they are first visited by \gls{DFS}. A \gls{DFS} postorder is a list of nodes that are in the order of when they are last visited by \gls{DFS}.

\subsubsection{Breadth-first Search}
\gls{BFS} is a graph traversal method that starts at source nodes (input nodes of the computation graph) and explores the graph by first visiting all the sibling nodes of the current nodes before moving to children nodes. A \gls{BFS} order of a graph is an enumeration of its nodes that is one possible output of applying \gls{BFS} on the graph.

\subsubsection{Lexicographical}
Lexicographical order is an order where the strings are placed in order based on the position of each character in the string and their position in the alphabet. For example, given the name strings of two nodes in a computation graph $a=a_1 a_2 \cdots a_k$ and $b=b_1 b_2 \cdots b_k$, the order of the name of the two nodes depends on the alphabetical order of the characters in the first place $i$ that $a$ and $b$ differs. If $a_i < b_i$ then $a < b$, otherwise $a > b$. For more concrete examples, see Appendix~\ref{appendix:sorting_order}.

\section{Graph Traversal Order in Device Placement}
In this section, we discuss challenges in device placement and impact of graph traversal order.

\subsection{Challenges in Device Placement}
Finding a good placement for model parallelization is challenging. Most of the state-of-the-art methods use \gls{RL} to find placements; however, \gls{RL} agents still require a long training time before they can find suitable placements. Mirhoseini et al.~\cite{mirhoseini2017device} find that it takes 12 to 27 hours for their \gls{RL} method to find the best placement. Although lots of efforts have been made in reducing the complexity of the problem~\cite{mirhoseini2018hierarchical}, making the training method more efficient~\cite{gao2018post,gao2018spotlight}, and generalizing them better on unseen computation graph~\cite{addanki2019placeto,zhou2019gdp,zhou2020single}, the training time of the \gls{RL} agent still remains long. 

One of the challenges in device placement is defining order for nodes in the computation graph $G$. Unlike text and image data, the nodes in graphs reside in a multi-dimensional space that are linked by edges to represent connectivity~\cite{ying2021transformers}. One has to transform graph data from the multi-dimensional space into a sequence of nodes before the majority of the \gls{DL} methods can consume the graph data. 
In Placeto~\cite{addanki2019placeto}, the structural information can be (partially) reflected in the sequential order that the auto device placement method iterates through the nodes of the computation graph.
Recent work in graph representation learning~\cite{ying2021transformers} has shown that successfully learning structural information of the graph helps better represent the graph. Better representations, in turn, lead to performance improvement of downstream tasks that utilize graph representations.

Another challenge in device placement concern the expressiveness of \gls{GNN} that are used to generate node embeddings. The \gls{GNN} that are used by state-of-the-art device placement methods mostly follow the message-passing paradigm, which is known to have inherent limitations. For example, the expressiveness of such \gls{GNN} is bounded by the Weisfeiler-Lehman isomorphism hierarchy~\cite{kreuzer2021rethinking}. Also, \gls{GNN}s are known to suffer from over-squashing~\cite{topping2021understanding}, where there is a distortion of information propagation between distant nodes. Due to these limitations, the node embeddings created by \gls{GNN} have limited expressiveness. In such cases, different graph traversal order in device placement can lead to placement with different performances.

\subsection{Impact of Graph Traversal Order}
Graph traversal order determines the order, which an \gls{RL} agent learns the placement of each node in the computation graph. We believe that the learning process of \gls{RL} agents for device placement can be improved if proper graph traversal order can be identified and used in the \gls{RL} training. Better placements could be found if an ordering of nodes can make the \gls{RL} agent prioritize the placement learning of important nodes that have a more significant impact on the placement execution time. For example, it might be easier for the \gls{RL} agent to first learn how to place the nodes that have heavy communications. On the other hand, misplacement of such nodes can lead to slower placement execution time due to extra data communication between different devices.

Order of placement in a local neighborhood could also play an important role. For example, modern \gls{DL} models are usually constructed using several computation blocks~\cite{szegedy2015going, vaswani2017attention, hochreiter1997long}. The input of the computation block might be copied and sent to a few parallel threads that perform computation independently. All the intermediate results from these parallel threads are later concatenated that will serve as the input of the next computation block. 
Suppose an \gls{RL} agent can not anticipate the concatenation of results from parallel computation threads. In that case, it might misplace the threads so that data communication becomes a bottleneck for the concatenation node. However, suppose the \gls{RL} agent first learns and decides on the placement of the concatenation node. In that case, it can better decide on the placement for the earlier node in the computation block to a better balance between computation and communication. We empirically study different graph traversal order mentioned in Section~\ref{sec:preliminaries:graph_traversal_order}.

\section{Evaluation}
This section presents the details of the empirical evaluation setup, results, experiment analysis and discussion, and guidelines for choosing graph traversal order for a given \gls{DL} model.

\subsection{Datasets}
We conduct our experiments on three different datasets \cifar, \nmt, and \ptb as in previous work~\cite{addanki2019placeto,mitropolitsky2020graph}. \cifar and \ptb are generated using an \gls{RL}-based method ENAS~\cite{pham2018efficient} that finds the optimal subgraph within a larger graph search space. The \cifar dataset consists of $32$ computation graphs of convolutional neural networks for image classification tasks. The \ptb dataset consists of $32$ computation graphs for language modeling tasks. The \nmt dataset contains $32$ variations of \gls{nmt}~\cite{wu2016google} with different number of unrolled steps. The computation graphs in \nmt are a family of encoder-decoder networks with attention structures. The nodes of computation graphs are pre-grouped together in all three datasets to reduce graph sizes in the same way as in~\cite{mirhoseini2017device}. The computation graphs in \cifar, \ptb, and \nmt have on average 300, 500, and 190 nodes. Table~\ref{tab:dataset_summary} summarizes the three datasets.

\subsection{Experiment Setup}

We implement all the graph traversal order in section~\ref{sec:preliminaries:graph_traversal_order} using NetworkX~\cite{hagberg2008exploring} and refer to them as \topo, \reversedtopo, \dfspre, \dfspost, \bfs, and \lexi hereafter. For the implementation of Placeto, we use the implementation provided in~\cite{mitropolitsky2020graph}, which is based on the original implementation~\cite{addanki2019placeto}. We use the same simulator in the original implementation to simulate the physical execution environment with different numbers of devices that a neural network can be placed on. We only change the graph traversal order of the \gls{RL} agent in each episode of training, and this order is fixed across different episodes that happened in one experiment.

We conduct experiments on the graphs from each of the three datasets with three, five, and eight devices, in line with previous work~\cite{mitropolitsky2020graph}. We run independent experiments with the same setting (dataset and number of devices) to account for the stochastic and randomness that might lead to differences in experiment results. We compare three different settings for the number of repeated runs on a subset of the whole datasets and found that $10$ repeated runs offer a good balance between computation load and the reproducibility of the result.

The experiments are run a standalone benchmark machine with AMD Ryzen Threadripper 2920X 12-Core Processor and 128 GB of RAM. Since we have $3 \times 32 \times 4 \times 6 \times 10 = 17280$ experiments to run, we use parallel docker containers that each have one experiment to speed up the experiment. We empirically found that the metrics we measure in the experiment are not sensitive to the number of parallel docker containers running simultaneously. We use TensorFlow and NetworkX libraries for the experiment, and we refer the readers to this repository\footnote{https://github.com/bwhub/Graph\_Traversal\_Order\_in\_Device\_Placement} for experiment code and the specific version of the libraries and other software settings.

\subsection{Results and Analysis}
Through the training process of device placement, an \gls{RL} agent aims to find device placement with lower execution times. Device placement training processes using different graph traversal order might have different learning speeds. Given the same amount of training time, an efficient graph traversal order finds a placement with lower execution times for the given input \gls{DL} model compared to the placements found by less efficient graph traversal order. We compare different graph traversal order based on the best placement execution time at episode $9$, $19$, and $49$. 

We empirically observe that the training process of the \gls{RL} agent can be roughly divided into three phases. In the first phase (episode $1$ to $9$), the RL agent learns efficiently and can find a better placement across different training episodes. This can be explained by the fact that the learning process just started, and finding a good enough placement that is better than a random strategy is not very hard. In the second phase (episode $10$ to $19$), the learning process slows down, and the \gls{RL} agent cannot always find drastically better placements than the first phase. This reflects that the learning process plateaus, and we see diminishing returns. In the third phase (episode $20$ to $49$), the \gls{RL} agent overcomes the plateau and finds better placement thanks to the more extended training budget and the knowledge learning through the process.

We report the number of times each graph traversal order finds placement with the lowest execution time for the input DL models in the given dataset. Each execution time is based on an average of $10$ repeated experiments to minimize the effect of randomness and stochastic effects during the training process. Table~\ref{tab:cifar10_best_order_count} shows the result of experiments on \cifar dataset. Although graph traversal order like \topo and \dfspre are the ones that find the placement with the lowest execution time, most of the time \reversedtopo and \dfspost are the best traversal order to use. This can be explained by the fact that there are structures of parallel convolutions in the computation graph of \cifar dataset where the intermediate results for parallel convolutions are concatenated for later use. In such cases, it is better to start the learning process from the nodes in the output layer of the model. Once the placement of concatenation nodes is settled, it will be easier for the \gls{RL} agent to optimize the placement for the parallel convolutions. Also, we observe that with more training episodes, \topo and \dfspre start to show fewer advantages as the number of times they find the best placement with the lowest execution time decreases.

We also find that the diameter of the computation graph for the \gls{DL} model also affects which graph traversal order is performing the best in the \cifar dataset. With a shorter diameter (e.g., diameter smaller than $100$), the \gls{DFS} family (\dfspre and \dfspost) performs the best. With a longer diameter (e.g., diameter larger than $100$), the topo family (\topo, \reversedtopo) tends to find better placements. This could be explained by the fact that the \gls{DFS} family forms longer sequences of consecutive nodes on the diameter with a larger diameter. This can be hard for the \gls{RL} agent to learn the placement of sibling nodes in the computation graph as they are far away from each other in the sequence. This might require the \gls{RL} agent to learn placement collocation of sibling nodes far away from each other.

Table~\ref{tab:nmt_best_order_count} shows the result of experiments on \nmt dataset. \reversedtopo order dominates and gives the best result. The can be explained by the fact that \reversedtopo order considers how intermediate results are concatenated in the computation graph. The \gls{RL} agent can decide the placement of the concatenation operation first. Then it is easier for the \gls{RL} agent to colocate the input operations nodes to the concatenation node to minimize expensive data transfer and synchronization between devices during training. In such cases, starting from the nodes in the output layers of the computation graph also helps.
\dfspost order does not work well on \nmt dataset as it has a larger average node degree of $2.65$ compared to the average node degree of $1.47$ of \cifar. This increases the effort for the \gls{RL} agent to collocate the sibling nodes that are far away in the placement sequence generated using \dfspost order. Better collocation of sibling nodes can also potentially explain why \bfs order is the graph traversal order that finds the placement with the lowest execution time.

Table~\ref{tab:ptb_best_order_count} shows the result of experiments on \ptb dataset. \bfs order is the graph traversal order that achieves the best learning efficiency. This can be explained by the fact that computation graphs in \ptb dataset have more nodes and edges than \cifar and \nmt datasets. There are potentially more sibling nodes that the \gls{RL} agent needs to consider when performing the placement. Since sibling nodes in a local neighborhood will be put close together in the traversal sequence generated by \bfs, it is easier for the \gls{RL} agent to learn to collocate these nodes together to avoid unnecessary data transfer between devices. In this way, the \gls{RL} agent does not need to worry too much about long-range dependencies in large computation graphs.

\begin{table}
\centering
\caption{Computation Graph Dataset Summary}
\label{tab:dataset_summary}
\begin{tabular}{rccc}
\hline
\multicolumn{1}{c}{\multirow{2}{*}{\textbf{Features}}} & \multicolumn{3}{c}{\textbf{Dataset}} \\ \cline{2-4} 
\multicolumn{1}{c}{} & cifar & nmt & ptb \\ \hline
\#nodes (avg) & 303.44 & 179.44 & \textbf{500.75} \\
\#edges (avg) & 444.22 & 476.25 & \textbf{1285.44} \\
node degree (avg) & 1.47 & \textbf{2.65} & 2.56 \\
diameter (avg) & 95.63 & 63.13 & \textbf{316.09} \\
diameter (min, max) & (74, 154) & (41, 69) & (216, 450) \\ \hline
\end{tabular}
\end{table}

\begin{table}
\centering
\caption{Best Traversal Order on \cifar dataset.}
\label{tab:cifar10_best_order_count}
\begin{tabular}{c|c|cccccc} 
\hline
\multicolumn{2}{c}{\textbf{cifar10 }} & \multicolumn{6}{c}{\textbf{Graph Traversal Order}} \\ 
\hline
\multicolumn{1}{c}{\textbf{\#dev}} & \multicolumn{1}{c}{\textbf{ep}} & \textbf{lexico} & \textbf{topo} & \textbf{dfs\_pre} & \textbf{rev\_topo} & \textbf{dfs\_post} & \textbf{bfs} \\ 
\hline
\multirow{3}{*}{\textbf{3dev}} & 9 & 1 & 6 & \textbf{10 } & \textbf{10 } & 4 & 1 \\
 & 19 & 3 & 3 & 7 & 8 & \textbf{11 } & 0 \\
 & 49 & 5 & 4 & 5 & \textbf{8 } & \textbf{8 } & 2 \\ 
\hline
\multirow{3}{*}{\textbf{5dev}} & 9 & 0 & \textbf{9} & 6 & 6 & \textbf{9 } & 2 \\
 & 19 & 0 & 9 & 3 & \textbf{10 } & 8 & 2 \\
 & 49 & 2 & 6 & 6 & \textbf{9 } & 6 & 3 \\ 
\hline
\multirow{3}{*}{\textbf{8dev}} & 9 & 0 & 8 & 3 & 10 & \textbf{11} & 0  \\
 & 19 & 1 & 4 & 2 & 8 & \textbf{17} & 0 \\
 & 49 & 0 & 5 & 1 & 11 & \textbf{15} & 0 \\
\hline
\end{tabular}
\end{table}

\begin{table}
\centering
\caption{Best Traversal Order on \nmt dataset.}\label{tab:nmt_best_order_count}
\begin{tabular}{c|c|cccccc} 
\hline
\multicolumn{2}{c}{\textbf{nmt}} & \multicolumn{6}{c}{\textbf{Graph Traversal Order}} \\ 
\hline
\multicolumn{1}{c}{\textbf{\#dev}} & \multicolumn{1}{c}{\textbf{ep}} & \textbf{lexico} & \textbf{topo} & \textbf{dfs\_pre} & \textbf{rev\_topo} & \textbf{dfs\_post} & \textbf{bfs} \\ 
\hline
\multirow{3}{*}{\textbf{3dev}} & 9 & 0 & 2 & 0 & \textbf{22} & 1 & 7 \\
 & 19 & 0 & 0 & 1 & \textbf{23} & 1 & 7 \\
 & 49 & 0 & 0 & 1 & \textbf{24} & 1 & 6 \\ 
\hline
\multirow{3}{*}{\textbf{5dev}} & 9 & 0 & 0 & 1 & \textbf{24} & 1 & 6 \\
 & 19 & 0 & 0 & 2 & \textbf{26} & 0 & 4 \\
 & 49 & 0 & 0 & 0 & \textbf{27} & 0 & 5 \\ 
\hline
\multirow{3}{*}{\textbf{8dev}} & 9 & 0 & 0 & 0 & \textbf{23} & 3 & 6 \\
 & 19 & 0 & 0 & 0 & \textbf{24} & 1 & 7 \\
 & 49 & 0 & 0 & 0 & \textbf{24} & 3 & 5 \\
\hline
\end{tabular}
\end{table}

\begin{table}
\centering
\caption{Best Traversal Order on \ptb dataset.}\label{tab:ptb_best_order_count}
\begin{tabular}{c|c|cccccc} 
\hline
\multicolumn{2}{c}{\textbf{ptb}} & \multicolumn{6}{c}{\textbf{Graph Traversal Order}} \\ 
\hline
\multicolumn{1}{c}{\textbf{\#dev}} & \multicolumn{1}{c}{\textbf{ep}} & \textbf{lexico} & \textbf{topo} & \textbf{dfs\_pre} & \textbf{rev\_topo} & \textbf{dfs\_post} & \textbf{bfs} \\ 
\hline
\multirow{3}{*}{\textbf{3dev}} & 9 & 0 & 1 & 6 & 1 & 0 & \textbf{24} \\
 & 19 & 0 & 1 & 10 & 1 & 2 & \textbf{18} \\
 & 49 & 0 & 1 & 8 & 5 & 2 & \textbf{16} \\ 
\hline
\multirow{3}{*}{\textbf{5dev}} & 9 & 0 & 0 & 3 & 0 & 2 & \textbf{27} \\
 & 19 & 0 & 1 & 3 & 2 & 2 & \textbf{24} \\
 & 49 & 0 & 1 & 2 & 6 & 5 & \textbf{18} \\ 
\hline
\multirow{3}{*}{\textbf{8dev}} & 9 & 0 & 0 & 2 & 1 & 1 & \textbf{28} \\
 & 19 & 0 & 0 & 2 & 5 & 4 & \textbf{21} \\
 & 49 & 0 & 0 & 2 & \textbf{13} & 5 & 12 \\
\hline
\end{tabular}
\end{table}

\subsection{Discussion and Guidelines}
In the previous subsection, we show that graph traversal order affects the training efficiency of the \gls{RL} agent, i.e., the execution time of the best placement found given the same amount of training budget. The optimal graph traversal order for the \gls{RL} agent depends on the characteristic of the neural network, e.g., number of nodes, average degree, that we want to find optimal placement on. Our findings are in line with previous research findings~\cite{mirhoseini2018hierarchical, addanki2019placeto} that graph traversal order does not affect the quality of the final placement found if ample training budget is given for finding that placement.

We have observed that given enough training budget, the difference of execution time between the placement found is less and less obvious since the \gls{RL} agent has managed to learn how to give good placement given enough training budget. However, our findings are still meaningful in real-world experiments where one cannot guarantee that the \gls{RL} agent could have an unlimited amount of time training. Under a limited training budget, a good graph traversal order could help to find better placement than other traversal order. Better placement improves the training throughput of the \gls{DL} model on distributed hardware. As a result of the better placement, one either choose to improve the training speed of the \gls{DL} model on the same dataset or train the \gls{DL} model on larger datasets within the same amount of time.

Identifying the proper graph traversal order for computation graph of \gls{DL} models can improve the training efficiency that leads to better placement with lower execution time on distributed hardware. However, finding the optimal graph traversal order for a given \gls{DL} model is not an easy task as many factors are involved in the process, e.g., the topology of the computation graph of the \gls{DL} model, the ratio of computation and communication during training. Although one cannot always quickly find the best graph traversal order for the computation of a given \gls{DL} model, we can still provide some guidelines based on our experience.

In general, it is good to start with graph traversal order that traverses the nodes in the computation graph in a backward fashion, i.e., start from the nodes in the final layer of the graph, gradually go through the nodes in the previous layers, and finish with the nodes in the first layer of the model. For example, when using \reversedtopo order, the \gls{RL} agent in the device placement method can first learn the placement of the nodes in the last layers and then on the nodes that are input to nodes that the \gls{RL} agent already find placement for. By starting from backward, the \gls{RL} agent can learn to better collocate parent and children nodes.
If the \gls{DL} model computation graph has a large diameter and a large number of nodes or groups of nodes, then graph traversal order that can put sibling nodes near each other in the one-dimensional sequence are better candidates for the optimal graph traversal order. For example, when facing a large \gls{DL} model with more than $200$ nodes, \bfs order can put sibling nodes close to each other in the one-dimensional sequence. Thus, the \gls{RL} agent can learn to better place the sibling nodes consecutively, instead of having to remember the placement of sibling nodes that are far away from each other in a long sequence.

In the context of ExtremeEarth project~\cite{koubarakis2019copernicus, koubarakis2021artificial, hagos2021extremeearth}, different types of models are used to provide \gls{EO} products. While hyperparameter tuning~\cite{meister2020maggy} and ablation studies~\cite{sheikholeslami2021autoablation} can help to improve model performance, identifying proper graph traversal order can improve the model parallel training performance.
For example, for \gls{SAR} image classification~\cite{khaleghian2021synthetic, khaleghian2021sea}, \reversedtopo and \dfspost would be good traversal order to start the experiment, as the models are similar to that in \cifar model dataset. For sequence classification tasks~\cite{paris2020monitoring}, \bfs would be a good traversal order to start with, as they are sequence to sequence models, which are similar to those in \ptb model datasets. \bfs order can help the \gls{RL} agent to collocate better the placement of sibling operations in the \gls{DL} model.

\section{Related Work}
In this section, we discuss related work on graph traversal order.

The previous work study relationship between graph traversal order and the execution time of the final placement found by an auto device placement method given enough training budget. HDP~\cite{mirhoseini2018hierarchical} randomized the order of groups on NMT (4-layer) baseline to feed into the Placer that predicts the placement for each group of nodes. The authors find that the difference between the fastest and slowest placements was less than $7\%$ in $10$ experiments.
Placeto~\cite{addanki2019placeto} uses \gls{GNN} to eliminate the need to assign indices when embedding graph features. Experiment results showed that the predicted placement of Placeto is more robust to graph traversal order than the RNN-based approaches.

REGAL~\cite{Paliwal2020Reinforced} uses topological ordering to convert a graph into a sequence. Mitropolitsky et al.~\cite{mitropolitsky2020graph} study how different graph embedding techniques affect the execution time of the final placement and show that position-aware graph embedding improves the execution time of the placement found compared to Placeto-GNN~\cite{addanki2019placeto} and GraphSAGE~\cite{hamilton2017inductive}. GPD~\cite{zhou2019gdp} removes the positional embedding in the transformer model to prevent overfitting.

Some work in other domains also studies graph traversal order. In chip placement, Mirhoseini et al.~\cite{mirhoseini2021graph} find that topological ordering can help the RL agent to place connected nodes close to each other. In the domain of generating graphs with \gls{DL} models, GraphRNN~\cite{you2018graphrnn} uses BFS order for graph generation to reduce the complexity of learning over all possible node sequences. The only possible edges for a new node are those connecting to nodes in the ``frontier'' of the \gls{BFS} order.

To the best of our knowledge, our work is the first to study how graph traversal order affects device placement training efficiency in device placement.


\section{Conclusion}

In this work, we study the impact of graph traversal order in device placement. We empirically show that different graph traversal order affect the learning efficiency of the auto device placement training process. An RL agent can learn more efficiently during the training process by finding placement strategies with lower execution time faster when given a proper graph traversal order. Specifically, we find that traversing the computation graph from the nodes in the output layer to the nodes in the input layer helps the RL agent find good placement efficiently in many cases. We also find that when an RL agent finds placement for larger computation graphs, traversing order that can better collocate sibling nodes, e.g., \gls{BFS}, in the traversal sequence is more efficient than its depth-first counterparts.

We provide practical guidelines on choosing the traversal order for device placement. We believe that our study can help researchers and practitioners better understand the relationship between types of network and graph traversal order. And the knowledge learned about traversal order can further generalize to learning settings for parallelization beyond device placement (model parallelization). 

There are several potential extensions and improvements, such as jointly learning graph traversal order, graph embedding, and the policy network in the RL agent. Another possible direction is to study graph traversal order based on the graph structures and features of individual nodes (e.g., input and output size, and computation intensity of the given node). Also, it would be interesting to see how the knowledge we learn on graph traversal order can be applied when using transformer for graphs for device placement.


\begin{appendices}
\section{nmt 64-30: node names in lexicographic order}
\label{appendix:sorting_order}

\begin{center}
\begin{tabular}{ |r|l| } 
\hline
Index & Node names \\ 
\hline
0 & decoder/attention\_decoder/attn\_0/concat\\
1 & decoder/attention\_decoder/attn\_1/concat\\
2 & decoder/attention\_decoder/attn\_10/concat\\
$\vdots$ & $\vdots$ \\
33 & decoder/depth\_0/static\_rnn\_0/add\_1\\
34 & decoder/depth\_0/static\_rnn\_1/add\_1\\
35 & decoder/depth\_0/static\_rnn\_10/add\_1\\
$\vdots$ & $\vdots$ \\
161 & encoder/attention/Reshape\_1\\
162 & encoder/depth\_0/lstm\_cell/bias/read\\
163 & encoder/depth\_0/static\_rnn\_0/add\_1\\
164 & encoder/depth\_0/static\_rnn\_1/add\_1\\
165 & encoder/depth\_0/static\_rnn\_10/add\_1\\
$\vdots$ & $\vdots$ \\
225 & encoder/slicing\_layer/strided\_slice\\
226 & init\_vars/global\_epoch\_step/Assign\\
227 & placeholders/concat\_1\\
\hline
\end{tabular}
\end{center}

\end{appendices}


\bibliographystyle{reference/IEEEtran}
\bibliography{reference/arXiv_submission.bib}

\end{document}